\PassOptionsToPackage{full}{textcomp}
\PassOptionsToPackage{usenames,svgnames,dvipsnames}{xcolor}
\documentclass[11pt,a4paper]{article}

\usepackage{times}
\usepackage{latexsym}
\usepackage{fontawesome}
\usepackage{microtype}
\usepackage{textcomp}
\usepackage[hyperref]{acl2021}

\usepackage{amsmath,graphicx}
\usepackage{algorithm}
\usepackage{algpseudocode}
\usepackage{url}
\usepackage{multirow}
\usepackage{CJKutf8}
\usepackage{comment}
\usepackage{adjustbox}
\usepackage{stix}
\usepackage{stfloats}
\usepackage{xspace}
\usepackage{listings}
\usepackage{booktabs}
\usepackage{adjustbox}
\usepackage{enumitem}
\usepackage[multiple]{footmisc}
\usepackage{appendix}
\usepackage{cleveref}

\aclfinalcopy % Uncomment this line for the final submission
\def\aclpaperid{3409} %  Enter the acl Paper ID here

\newcommand{\systemname}{\mbox{KaggleDBQA}\xspace}
\newcommand{\pkey}{{\tiny\faKey}\xspace}
\newcommand{\eg}{\textit{e.g.}\xspace}
\newcommand{\ie}{\textit{i.e.}\xspace}

\newcommand{\sql}[1]{\lstinline[language=SQL, basicstyle=\footnotesize\ttfamily]{#1}}

\title{{KaggleDBQA}: {R}ealistic {E}valuation of {T}ext-to-{SQL} {P}arsers}

\newcommand\uw{$^{\diamondsuit}$}
\newcommand\ms{$^\spadesuit$}
\newcommand\aspace{\hspace{.75em}}

\author{
Chia-Hsuan Lee\uw\aspace
Oleksandr Polozov\ms\aspace
Matthew Richardson\ms\aspace \\
\uw University of Washington \aspace\ms Microsoft Research, Redmond \\
{\tt chiahlee@uw.edu}\\
{\tt \{polozov,mattri\}@microsoft.com}
}

\begin{document}
\maketitle
\begin{abstract}
The goal of database question answering is to enable natural language querying of real-life relational databases in
diverse application domains.
Recently, large-scale datasets such as Spider and WikiSQL facilitated novel modeling techniques for text-to-SQL parsing,
improving zero-shot generalization to unseen databases.
In this work, we examine the challenges that still prevent these techniques from practical deployment.
First, we present \systemname, a new cross-domain evaluation dataset of real Web databases, with domain-specific data
types, original formatting, and unrestricted questions.
Second, we re-examine the choice of evaluation tasks for text-to-SQL parsers as applied in real-life settings.
Finally, we augment our in-domain evaluation task with \emph{database documentation}, a naturally occurring source of
implicit domain knowledge.
We show that \systemname presents a challenge to state-of-the-art zero-shot parsers but a more realistic evaluation
setting and creative use of associated database documentation boosts their accuracy by over 13.2\%, doubling their
performance.
\end{abstract}

\section{Introduction}
\label{sec:intro}

\begin{figure*}[t]
    \small
    \textbf{Database:} \sql{Student Math Score} \\[5pt]
    \setlength{\tabcolsep}{4pt}
    \textbf{Table \sql{FINREV_FED_17}: }
    \adjustbox{valign=t}{\begin{tabular}{|l|p{2.9cm}|l|l|l|l|l|}
            \hline
            \pkey\sql{state_code} & \sql{school_district} & \sql{yr_data} & \sql{t_fed_rev} & \sql{c14} & \sql{c15} & \vdots \\
            \hline
            33 & NEW YORK CITY SCHOOL DISTRICT & 17 & 2061297 & 956851 & 439209 & \vdots \\
            47 & FAIRFAX CO SCHS & 17 & 126916 & 21035 & 36886 & \vdots \\
            \hline
    \end{tabular}} \\[1pt]
    \textbf{Column Descriptions: }
    \adjustbox{valign=t}{\begin{tabular}{ll}
            \sql{t_fed_rev} & Total federal revenue through the state to each school district \\
            \sql{c14} & Federal revenue through the state-Title 1 (no child left behind act) \\
            \sql{c15} & Federal revenue through the state - Child Nutrition A \\
    \end{tabular}}
    \\[5pt]
    \textbf{Table \sql{FINREV_FED_17_KEY}: }
    \adjustbox{valign=t}{\begin{tabular}{|l|l|l|}
            \hline
            \pkey\sql{state_code} & \sql{state} & \sql{#_Records} \\
            \hline
            1 & Alabama & 137 \\
            $\cdots$ & $\cdots$ & $\cdots$ \\
            50 & Wisconsin & 425 \\
            51 & Wyoming & 48 \\
            \hline
    \end{tabular}}
    \\[4pt] \noindent\textcolor{gray}{\rule{\textwidth}{0.5pt}} \\[4pt]
    \begin{tabular}{lp{14cm}}
        \hspace{-7pt} \textbf{Example Question:} &
        \textit{Which school district received the most of federal revenue through state in Wisconsin?} \\
        \hspace{-7pt} \textbf{Example SQL:} &
        \vspace{-2\medskipamount}
        \begin{lstlisting}[language=SQL,gobble=12]
            SELECT T1.school_district
            FROM FINREV_FED_17 as T1 JOIN FINREV_FED_KEY_17 as T2
            ON T1.state_code = T2.state_code WHERE T2.state = "Wisconsin"
            ORDER BY T1.t_fed_rev DESC LIMIT 1
        \end{lstlisting}
        \vspace{-3\medskipamount}
    \end{tabular}
    \caption{Two table excerpts from the Student Math Score database in \systemname and an example
        question-SQL pair.
        The column names are abbreviated (\eg \sql{t_fed_rev}) or obscure (\eg \sql{c14}, \sql{c25}) but
        documentation (\eg column descriptions) alleviates this.
        Source: \url{https://kaggle.com/loganhenslee/studentmathscores}.
        % Our dataset contains complex queries that require understanding multiple tables.
    }
    \label{fig:example}
\end{figure*}

Text-to-SQL parsing is a form of \emph{database question answering} (DBQA) that answers a user's natural-language
(NL) question by converting it into a SQL query over a given relational database.
It can facilitate NL-based interfaces for arbitrary end-user applications, thereby removing the need
for domain-specific UX or learning query languages.
As such, DBQA attracted significant attention in academia and industry, with development of supervised
datasets~\cite{yu-etal-2018-spider}, large-scale models~\cite{wang-etal-2020-rat,zeng-etal-2020-photon}, and novel modeling
techniques~\cite{yu2020grappa,deng2020structure}.

The key challenge of text-to-SQL parsing is \emph{zero-shot generalization} to unseen domains, \ie
to new database schemas and differently distributed NL questions.
Large-scale annotated datasets like Spider~\cite{yu-etal-2018-spider} and WikiSQL~\cite{zhong2017seq2sql}
evaluate \emph{cross-domain generalization} of text-to-SQL parsers by restricting overlap between train and test
domains.
Such challenging benchmarks facilitate rapid progress in DBQA.
State-of-the-art (SOTA) accuracy on Spider rose from 12.4\% to 70.5\% in just two years since its release,
demonstrating the value of well-chosen evaluation settings.

Despite impressive progress in DBQA, deployment of SOTA parsers is still challenging.
They often lack robustness necessary to deploy on real-life application domains.
While many challenges underlie the gap between SOTA DBQA and its real-life deployment, we identify three
specific discrepancies. %between the academic and industrial DBQA evaluation settings.

% However, the evaluation setting of Spider/WikiSQL is still disconnected from real-world conditions.
% it simplifies zero-shot generalization to industrial settings by normalizing and preprocessing
First, Spider and WikiSQL datasets normalize and preprocess database schemas or rely on academic example databases that
originate with human-readable schemas~\cite{suhr2020exploring}.
In contrast, industrial databases feature abbreviated and obscure naming of table, columns, and data values, often
accrued from legacy development or migrations.
\Cref{fig:example} shows a characteristic example.
After deployment, text-to-SQL parsers struggle with \emph{schema linking} to domain-specific entities because
they do not match the distribution seen in their pre-training (\eg BERT) or supervised training (\eg Spider).

Second, the NL questions of Spider and WikiSQL have high \emph{column mention
percentage}~\cite{deng2020structure}, which makes their language unrealistic.
This can be an artifact of rule-generated NL templates (as in WikiSQL) or annotation UIs that prime the
annotators toward the schema (as in Spider).
Either way, real-world deployment of a text-to-SQL parser optimized on Spider faces a
distribution shift in NL, which reduces its realistic performance.

% Third, it assumes no in-domain supervision other than database schema. This is for good reasons -- challenging problem
% setup inspires scientific innovation. However, many sources of implicit knowledge can augment zero-shot generalization,
% and also it is never zero-shot in reality.
Finally, the standard evaluation setting of cross-domain text-to-SQL parsing assumes no in-domain supervision.
This simplifies parser evaluation and raises the challenge level for zero-shot generalization.
However, it does not leverage knowledge sources commonly present in real-world applications, both explicit (annotated
in-domain examples) and implicit (\eg database documentation, SQL queries in the application
codebase, or data distributions).
A well-chosen alternative evaluation setting would facilitate development of DBQA technologies that match their
real-world evaluation.

\paragraph{\systemname}
We introduce \systemname, a new dataset and evaluation setting for text-to-SQL parsers to bridge the gap between SOTA
DBQA research and its real-life deployment.\footnote{Available at \url{https://aka.ms/KaggleDBQA}.}
It systematically addresses three aforementioned challenges:
\begin{itemize}[left=0pt]
    \item To test database generalization, it includes real-world databases from
        Kaggle,\footnote{\url{https://www.kaggle.com}} a platform for data science competitions and dataset
        distribution.
        They feature abbreviated and obscure column names, domain-specific categorical values, and minimal
        preprocessing (\Cref{sec:dataset:kaggle}).
    \item To test question generalization, we collected unrestricted NL questions over the databases in \systemname.
        Importantly, the annotators were not presented with original column names, and given no task priming
        (\Cref{sec:dataset:questions}).
        Out of 400 collected questions, one-third were out of scope for SOTA text-to-SQL parsers.
        The remaining 272 questions, while expressible, can only be solved to 13.56\% accuracy (\Cref{sec:evaluation}).
    \item Finally, we augment \systemname with \emph{database documentation}, common metadata for real-world 
        databases and a rich source of implicit domain knowledge.
        Database documentation includes column and table descriptions, categorical value descriptions (known as
        \emph{data dictionaries}), SQL examples, and more (\Cref{sec:dataset:docs}).
        We present a technique to augment SOTA parsers with column and value descriptions, which significantly improves their
        out-of-domain accuracy (\Cref{sec:evaluation}).
\end{itemize}
\Cref{fig:example} shows a representative example from the dataset.
Aligning \emph{``federal revenue''} and \sql{t_fed_rev} is hard without domain knowledge.

In addition to more realistic data and questions, we argue that evaluation of real-world text-to-SQL performance should
assume \emph{few-shot} access to $\sim$10 in-domain question-SQL examples rather than measuring \emph{zero-shot} performance.
In practical terms, few-shot evaluation assumes up to 1-2 hours of effort by a target database administrator or
application developer, and translates to significant performance benefits.
In a few-shot evaluation setting, augmenting a SOTA text-to-SQL parser (RAT-SQL by \citet{wang-etal-2020-rat}) with
database documentation almost doubled its performance from 13.56\% to 26.77\%.
See \Cref{sec:evaluation}.

\section{Related Work}
\label{sec:related}

\paragraph{Text-to-SQL Semantic Parsing}
Semantic parsing has been studied extensively for decades~\cite{cacm-liang2016learning}.
Key \emph{in-domain} datasets such as GeoQuery~\cite{geoquery} and ATIS~\cite{atis} acted as initial catalyst for the 
field by providing an evaluation measure and a training set for learned models.
Applying a system to a domain with a different distribution of questions or parses required out-of-domain data or
domain transfer techniques.
Recently, \emph{cross-domain} datasets WikiSQL~\cite{zhong2017seq2sql} and
Spider~\cite{yu-etal-2018-spider} proposed a \emph{zero-shot} evaluation methodology that required out-of-domain
generalization to unseen database domains.
This inspired rapid development of \emph{domain-conditioned} parsers that work ``out of the box'' such as
RAT-SQL~\cite{wang-etal-2020-rat} and IRNet~\cite{irnet}. 
We use the same exact match accuracy metric as these works. Recent work \cite{zhong2020semantic} has proposed evaluating SQL prediction via semantic accuracy by computing denotation accuracy on automatically generated databases instead. 

\paragraph{Few-shot learning}
In this paper, we propose a \emph{few-shot} evaluation to inspire future research of practical text-to-SQL parsers.
Like zero-shot, few-shot has access to many out-of-domain examples, but it also has access to a small number of
in-domain examples as well.
Few-shot learning has been applied to text classification in ~\cite{mukherjee2020uncertainty}, and has also been applied to semantic parsing. Common techniques include meta-learning~\cite{huang2018natural,wang2020meta,li2021few,sun2020neural} and adversarial learning~\cite{li2020domain}. 

\paragraph{Generalization and Practical usability}
Recent work has begun to question whether existing datasets are constructed in a way that will lead to models that
generalize well to new domains.
\citet{suhr2020exploring} identified a number of challenges with text-to-SQL datasets, one of which is an artificially
high overlap between words in a question and words in the tables.
This issue appears in Spider and is a byproduct of the fact that question authors view the database schema as they write
their question.
The Spider-Realistic~\cite{deng2020structure} dataset aims to reduce this by explicitly rewriting the questions to avoid
overlapping terms.

Other works has studied the problem of the gap between academic datasets and their practical
usability~\cite{de2020towards,radhakrishnan2020colloql,zhang2020did}, including highlighting the need for data to be real.
Our goal was to create an evaluation dataset and metric that minimizes this gap; our dataset is constructed from real data
found on Kaggle that has been used for competitions or other analyses.

Another direction of generalization being explored is compositionality.
\citet{keysers2019measuring} used rules to generate a large-scale semantic parsing dataset that specifically tests models for
composability. 

\paragraph{Leveraging other resources for learning}
% Sometimes, additional resources are brought to bear on a task.
\citet{rastogi2020towards} provide NL descriptions for slots and intents to help dialogue state tracking.
\citet{logeswaran2019zero} use descriptions to facilitate zero-shot learning for entity linking.
\citet{weller2020learning} use descriptions to develop a system that can perform zero-shot learning on new tasks.
We follow by including documentation on each included real-world database.
Notably, this documentation was \emph{written for human consumption of the database} rather than prepared for
\systemname, and thus is a natural source of domain knowledge.
It provides similar benefits to codebase documentation and comments, which improve source code
encoding for AI-assisted software engineering tasks~\cite{panthaplackel2020associating,wei2019code}.

\section{\systemname: A Real World Dataset}
\label{sec:dataset}

The goal of the \systemname evaluation dataset is to more closely reflect the data and questions a text-to-SQL
parser might encounter in a real-world setting.
As such, it expands upon contemporary cross-domain text-to-SQL datasets in three key aspects:
\textbf{(i)} its \textbf{databases} are pulled from real-world data sources and \emph{not} normalized;
\textbf{(ii)} its \textbf{questions} are authored in environments that mimic natural question answering;
\textbf{(iii)} its \textbf{evaluation} assumes the type of system augmentation and tuning that could be expected from
domain experts that execute text-to-SQL parser deployment.
We describe each of these components in turn in this section.

\begin{table*}[t]
    \centering
    \small
    \caption{Comparison of text-to-SQL datasets.
        We follow the data filtering rules of \citet{suhr2020exploring} and \citet{deng2020structure}, which reduces the effective number of examples from the original datasets to make them consistent.
        \%WHERE measures the percentage of examples where all
        \sql{WHERE}/\sql{HAVING} columns in the SQL query are explicitly mentioned in the NL question.
        \%VAL compares all the values in the SQL queries; \%SELECT compares all the \sql{SELECT} columns;
        \%NON SELECT compares all columns except the \sql{SELECT} columns.
        \systemname has low column mention percentage and contains databases with multiple tables.
        % The SQL query structures are more complex than the Spider.
        % Please refer to figure~\ref{fig:hardness}.
    }
    \label{tab:stats}
    \begin{tabular}{lcccccccc}
    \toprule
        Dataset & \# Examples                 & \# DB  & \# Table/DB  & \% WHERE  &  \% VAL &  \% SELECT  & \% NON-SELECT  \\
        \midrule
                        ATIS & 275  & 1 & 25 & 0.0 & 95.6 & 0.0 & 0.0 \\
                        GeoQuery & 525 & 1 & 7 & 3.8 & 100.0 & 32.9 & 9.1 \\
                        Restaurants & 39 & 1  & 3 & 0.0 & 100.0 & 0.0 & 0.0 \\
                        Academic & 179 & 1  & 17 & 5.2 & 100.0 & 15.1 & 1.7 \\
                        IMDB & 111 & 1 & 17 & 1.6 & 100.0 & 7.1 & 0.8  \\
                        Yelp & 68  & 1  & 8 & 4.2 & 100.0 & 5.7 & 4.1 \\
                        Scholar & 396 & 1  & 10 & 0.0 & 100.0 & 0.7 & 0.2  \\
                        Advising & 281 & 1  & 15 & 4.0 & 100.0 & 6.1 & 3.9 \\
                        \midrule
                        %WikiSQL & 80654  &  26521 & - & 1 & ? \\
                        Spider Train &  7000 & 140 &  5.26 & 40.8  & 89.01  & 52.4 & 41.6  \\
                        Spider Dev & 1034  & 20 &  4.05  & 39.2 & 91  & 48.2 &  33.1 \\
                        %Spider Realistic &   &  &  &  &   & 1.8  & 93.27  &  54.3 &  5.3 \\
                        \midrule
                        \textbf{\systemname} & 272 & 8  &  2.25 & 8.7 & 73.5 & 24.6 & 6.8 \\
                        \bottomrule
    \end{tabular}
\end{table*}

\subsection{Database Collection}
\label{sec:dataset:kaggle}
We chose to obtain databases from Kaggle, a popular platform for hosting data
science competitions and sharing datasets and code.
Their hosted datasets are by definition ``real'' as they are used by members of the site for research.
Competition hosts upload their data unnormalized, and the data content and formatting matches its domain-specific usage
(see \Cref{fig:example} for an example).
% \matt{we need to strengthen our case that these are real, not toy data uploaded by members.
% Is there a way to see if the datasets were used by anyone else?
% Maybe we can look at the ones we ended up with and show they all come from databases originally or something?}
To construct \systemname, we randomly selected 8 Kaggle datasets that satisfied the following criteria:
(a) contained a SQLite database;
(b) licensed under a republishing-permissive license;
(c) had associated documentation that described the meaning of the tables and columns.

\subsection{Questions}
\label{sec:dataset:questions}
For each database, we asked five annotators to write ten domain-specific questions that they think someone might be
interested in and that can be answered using the database.
We use five annotators per database to help guarantee diversity of questions.
Each annotated two databases, for a total of 20 annotators and 400 questions.

The annotators are not required to possess SQL knowledge so their questions are more reflective of natural user
interests.
Importantly, to discourage users from using the same terms from the database schema in their questions, \emph{we replace
the original column names with the column descriptions}.
When annotating the questions, the annotators are shown a paragraph description of the database, table names, column
descriptions and ten sampled rows for each table.
We do not provide any constraints or templates other than asking them to avoid using exact phrases from the column
headings in the questions.
\Cref{sec:annotation} shows the full guidelines.

Separately, each question is annotated with its SQL equivalent by independent SQL experts.
They are given full access to all of the data content and database schema.
One-third of the questions were yes/no, percentage, temporal, or unexpressible in SQL and were not considered in our
evaluation of SOTA models (see \Cref{sec:question_types} for details), leaving 272 questions in total.

\subsection{Database Documentation}
\label{sec:dataset:docs}
Each database has associated plain-text \emph{documentation} that can assist text-to-SQL parsing.
It is commonly found as internal documentation for database administrators or external documentation accompanying a
dataset release.

The contents vary but often contain an overview of the database domain, descriptions of tables and columns, sample
queries, original sources, and more.

While all of these types of information could be leveraged to assist with domain transfer,
in this work we focus on the \emph{column descriptions}.
They help address the \emph{schema linking} problem of text-to-SQL parsing, \ie aligning entity references in the
question with database columns~\cite{wang-etal-2020-rat}.
For example, \textit{``federal revenue''} in \Cref{fig:example} must be aligned to the column \sql{t_fed_rev}
even though its abbreviated name makes alignment non-obvious.

We manually extract the column descriptions from the database documentation and provide the mapping from column to
description as part of \systemname.
The descriptions are free text and sometimes contain additional information such as defining the values in an categorical
column.
Such information could help with the \textit{value-linking} problem (mapping a value in the question to the column that
likely contains it).
We leave the entire description as a single field and leave it to future work to explore these uses further.
In addition to column descriptions, we also include the original unstructured documentation which can be used for future
research on automatically extracting descriptions or leveraging other domain knowledge.

\subsection{Few-shot Evaluation Setting}
The current cross-domain datasets Spider~\cite{yu-etal-2018-spider} and WikiSQL~\cite{zhong2017seq2sql} evaluate models
in a \textit{zero-shot} setting, meaning the model is trained on one set of domains and evaluated on a completely
disjoint set.
This evaluation encourages the development of systems that work well "out of the box" and has spurred great development
in cross-domain text-to-SQL systems that are able to generalize to new domains.
However, we believe the zero-shot setting is overly-restrictive compared to how text-to-SQL systems are likely to be
actually used in practice.

We postulate that it is more realistic to assume a setting where an application author spends 1-2 hours authoring
examples and adapting existing database documentation.
This time investment is a small fraction of the time required to prepare an application itself and so we believe
application authors would devote the time if it resulted in increased text-to-SQL accuracy.
In informal experiments, we have found SQL annotators can author 10-20 examples in an hour.
Thus, the \systemname evaluation setting is \textit{few-shot}: 30\% of the questions for each domain (6-15 depending
on the domain) are designated as \textit{in-domain} and may be used as part of training for that domain,
along with documentation.
The remaining 70\% are used for evaluation.

We report accuracy in both the \textit{few-shot} as well as the standard \textit{zero-shot} (cross-domain) setting in
this paper, but consider the few-shot setting to be the primary evaluation setting for \systemname.
% Indeed, we would encourage that it should also be a reported metric of other cross-domain datasets such as Spider.
% \matt{don't say this if we don't have results on Spider.}
Evaluation is conducted on the same 70\% portion regardless of setting, to ensure comparable results.

\begin{table}[t]
    \small
    \centering
    \caption{Average partial match \% of columns descriptions across examples. We check whether 1- to 3-grams in the question are part of any column descriptions.}
    \label{tab:description}
    \begin{tabular}{lccc}
    \toprule
        Type of n-gram  &  1 & 2 & 3 \\
        \midrule
        \% Cols matched in golden SQL  & 56.27 & 21.47 & 4.80  \\
         \# Cols matched in golden SQL  & 1.06 & 0.37 & 0.07  \\
        \# Cols matched not in the SQL & 4.69 & 1.29 & 0.13 \\
        \bottomrule
    \end{tabular}
\end{table}

\begin{figure}[t]
    \centering
    \includegraphics[width=\linewidth]{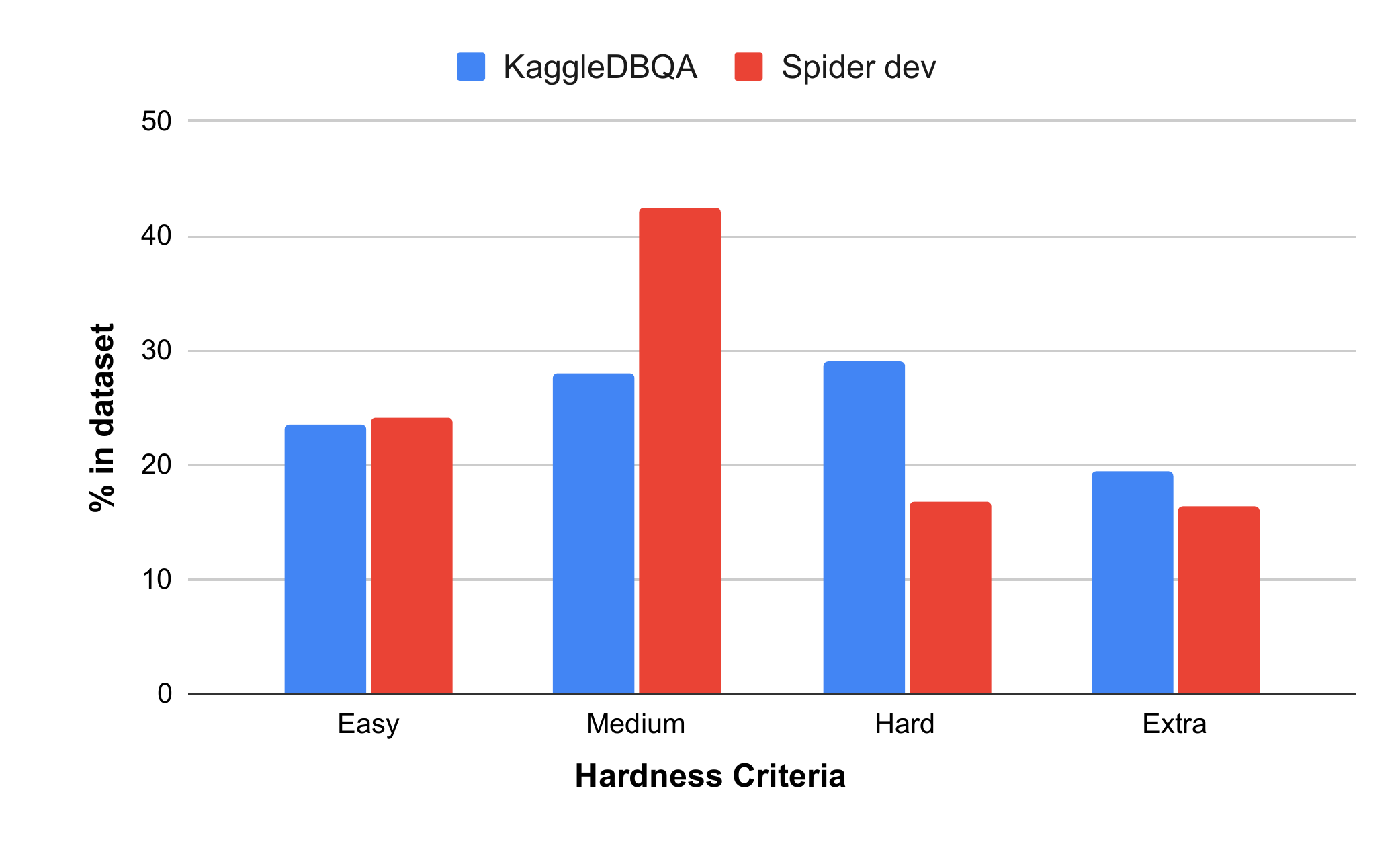}
    \caption{Comparisons of text-to-SQL datasets in terms of SQL structure hardness.
    \systemname has more complex SQL query structure than the Spider dev set.}
    \label{fig:hardness}
\end{figure}

\subsection{Dataset Statistics and Comparison}
We compare \systemname with previous benchmark datasets using key metrics in Table~\ref{tab:stats}.
\systemname has the lowest value mention percentage among all datasets, and also exhibits a low overlap between question
terms and column names similar to that in all of the datasets besides Spider, making it more in line with what would be
expected in a real-world setting where the people asking questions are not familiar with the actual database schema and
terminology. This is likely a result of replacing column names with descriptions in the question annotation task.

We also analyze the overlap between question terms and column descriptions in Table~\ref{tab:description}.
Because the descriptions are significantly longer than column names, we require only that they share an n-gram in common
(ignoring stop-words) rather than requiring exact match as was done for column mention percent.
Unigram overlap is reasonably high (56\% of correct columns match the question) but also results in many false-positive
matches with other columns.
Increasing n-gram size decreases false-positives but also rapidly decreases the correct column match percent.
Thus, column descriptions may help guide the model, but are not as strong of a signal as found in Spider which suffers
from high exact column name match overlap.
This was our intention in asking our annotators to avoid using the descriptions verbatim when writing questions.

To measure the complexity of SQL in \systemname, we adopt the hardness criteria of Spider and report the
numbers in~\Cref{fig:hardness}.
The queries are on average more complex than Spider's, with significantly more hard and extra-hard ones.

\section{Experiments}
\label{sec:evaluation}

\subsection{Baseline Results}
We first evaluate \systemname using models that were developed for the Spider dataset.

\paragraph{EditSQL~\cite{editsql}:}
EditSQL (with BERT) is the highest-performing model on the Spider dataset that also provides an open-source
implementation along with a downloadable trained model.\footnote{\url{https://github.com/ryanzhumich/editsql}}
The model was built for edit-based multi-turn parsing tasks, but can also be used as a single-turn parser for Spider
or \systemname.
It employs a sequence-to-sequence model with a question-table co-attention encoder for schema encoding.

\paragraph{RAT-SQL~\cite{wang-etal-2020-rat}:}
RAT-SQL (v3 + BERT) is the model with highest accuracy on the Spider leaderboard that also provides an open-source
implementation.\footnote{As of one month before paper authoring.
Current SOTA systems are also based on RAT-SQL and add less than 5\% accuracy, thus will likely behave similarly.}\footnote{\url{https://github.com/microsoft/rat-sql}}
It adds string matching to the encoder through the use of relation-aware self-attention and adopts a tree-based decoder
to ensure the correctness of the generated SQL.

Throughout this paper, we use the same \textit{exact-match accuracy} metric introduced by the Spider dataset.
Although our primary evaluation setting is few-shot, we first examine the traditional zero-shot setting to
present an unbiased comparison with previous results.
Table~\ref{tab:sota} compares the performance of these two models (both trained on Spider).
As can be seen, the performance of both models is significantly lower on \systemname.
This echoes the findings of \citet{suhr2020exploring} who found that a model trained on Spider did not generalize well to other datasets.
Also, \systemname has much fewer column mentions and much more complex SQL than Spider (see~\Cref{tab:stats} and~\Cref{fig:hardness}).

For all further experiments on \systemname that emulate real-world evaluation, we choose RAT-SQL as the best performing
parser.

\begin{table}[t]
    \setlength{\tabcolsep}{5pt}
    \centering
    \small
    \caption{Zero-shot testing results of various open-source models on \systemname and on the test set of Spider. All
        numbers are the exact match accuracy evaluated by the Spider official scripts. The Spider results are from the
        official leaderboard. The \systemname results are the average of three different runs.
    }
    \label{tab:sota}
    \begin{tabular}{lcc}
        \toprule
        \textbf{Models}  & \textbf{Spider} &  \textbf{\systemname} \\
        \midrule
        \textbf{RAT-SQL}~\cite{wang-etal-2020-rat}  & 65.60  & 13.56  \\
        \textbf{EditSQL}~\cite{editsql} &  53.40 & 11.73  \\
        \bottomrule
    \end{tabular}
\end{table}

\subsection{RAT-SQL on \systemname}

\subsubsection{Moving to the Few-Shot Setting}
To apply RAT-SQL to \systemname's few-shot setting, for each domain we create a model by fine-tuning on its 30\%
in-domain data.
See \Cref{sec:implementation_details} for implementation details.
This fine-tuning is always performed as the last step before evaluation. 

As~\Cref{tab:indomain} shows, fine-tuning on a small amount of in-domain data
dramatically increases overall accuracy from 13.56\% to 17.96\% (rows (a) and (e)),

Although the few-shot setting is our primary setting, we also present results in the zero-shot setting to compare to previous work (Table~\ref{tab:indomain} rows (e)-(h)). However, in the remainder of the paper we will be focusing on the few-shot setting.

\begin{table*}[t]
    \small
    \centering
    \caption{Exact match accuracy and standard error on \systemname, mean of three runs with different random seeds.}
        % All numbers are the exact match accuracy as defined by Spider evaluation metric.
    \label{tab:indomain}
    \vspace{-0.5\baselineskip}
    \adjustbox{max width=\textwidth}{
    \begin{tabular}{lccccccccc}
        \toprule
        \multicolumn{10}{c}{With \textit{fine-tuning}} \\
        \midrule
        Models  & Nuclear & Crime & Pesticide & MathScore & Baseball & Fires & WhatCD & Soccer & Avg  \\
        \midrule
        (a) RAT-SQL  & 28.78 & 35.18 & 11.76 & 3.50  & 14.81  & 30.66  & 10.68 & 8.33  &  $17.96  \pm 0.5\% $\\
        (b) \quad \textit{w.} desc & 22.72  &  29.62 &  12.74  &  3.50 & 11.11 & 33.33 & 19.04 & 8.33 & $17.55 \pm 0.6\%$\\
        (c) \quad \textit{w.} \textit{adaptation}  & 28.78  & 44.44 & 16.66 & 8.76 & 16.04 & 37.33 & 16.66 & 13.87 & $22.82 \pm 0.1\%$ \\
        (d) \quad \textit{w.} desc \textit{+} \textit{adaptation}  & 36.35 & 44.44 & 21.56 & 7.01 & 22.22 & 41.33 & 27.38 & 13.87 & \textbf{$26.77  \pm 0.4\%$} \\
        \midrule
        \multicolumn{10}{c}{Without \textit{fine-tuning}} \\
        \midrule
        Models  & Nuclear & Crime & Pesticide & MathScore & Baseball & Fires & WhatCD & Soccer & Avg  \\
        \midrule
        (e) RAT-SQL  & 22.72 & 25.92 &  8.82 & 0.00  & 12.34  & 17.33  & 4.76 & 16.66  & $13.56 \pm 0.1\% $\\
        (f) \quad \textit{w.} desc  & 24.24  & 20.37 & 7.84  & 0.00 & 9.87 & 13.33 & 7.14 & 16.66  & $12.43  \pm 0.1\% $\\
        (g) \quad \textit{w.} \textit{adaptation}   & 25.75  & 38.88 &  12.74 & 3.50 & 7.40 &  20.00 & 9.52 &  16.66 & $16.80 \pm 0.8\%$\\
        (h) \quad \textit{w.} desc \textit{+} \textit{adaptation}  & 30.29 &   25.92   &   17.64  &   3.50  & 16.04 &  25.33 &  11.9 &  16.66 &  $18.41  \pm 0.4\%$\\
        \bottomrule
    \end{tabular}}
\end{table*}

\begin{table}[b]
    \small
    \centering
    \caption{Exact match accuracy and standard error on schema-normalized \systemname, average of three runs with different random seeds.}
    \label{tab:normalized}
    \vspace{-0.5\baselineskip}
    \begin{tabular}{lc}
        \toprule
        \multicolumn{2}{c}{With  \textit{fine-tuning}} \\
        Models  & Avg  \\
        \midrule
        (a) RAT-SQL  &   $17.96  \pm 0.5\% $ \\
        (b) \quad \textit{w.} desc  &  $17.55  \pm 0.6\%$ \\
        (c) \quad \textit{w.} normalization  &  $23.09  \pm 0.9\%$\\
      
        (e) \quad \textit{w.} \textit{adaptation}   &  $22.82  \pm 0.1\%$ \\
        (f) \quad \textit{w.} desc \textit{+} \textit{adaptation}  & $26.77 \pm 0.4\%$\\
        (g) \quad \textit{w.} normalization \textit{+} \textit{adaptation}  & $25.60 \pm 0.9\%$ \\
        (h) \quad \textit{w.} desc \textit{+} normalization \textit{+} \textit{adaptation}  &  $27.83 \pm 0.7\%$ \\
        \bottomrule
    \end{tabular}
\end{table}

\begin{table*}[t]
    \small
    \centering
    \caption{
        Examples where description-augmented (``desc.'') models solve a question that unaugmented models (``no desc.'') do not.
        Both models are adapted and fine-tuned. Both omit values, as per the official Spider metric.
    }
    \label{tab:correct}
    \vspace{-0.5\baselineskip}
    \adjustbox{max width=\textwidth}{
    \begin{tabular}{lp{12.0cm}}
        %\hline
        %\textbf{Models}  &  {SQL Query} \\
        \toprule
        \textbf{Database \sql{USWildFires}} & \textbf{Column Descriptions}\\
        \sql{STAT\_CAUSE\_CODE}   & Code for the (statistical) cause of the fire  \\
        \sql{STAT\_CAUSE\_DESCR}   & Description of the (statistical) cause of the fire.  \\        
        \sql{FIRE\_SIZE}  & Estimate of acres within the final perimeter of the fire\\
        \midrule
        \textbf{Question}  & What’s the most common cause of the fire (code) in the database?\\
        \textbf{no desc.} & \sql{SELECT Fires.STAT\_CAUSE\_DESCR FROM Fires GROUP BY Fires.STAT\_CAUSE\_DESCR ORDER BY Count(*) DESC LIMIT 1} \\
        \textbf{desc.} & \sql{SELECT Fires.STAT\_CAUSE\_CODE FROM Fires GROUP BY Fires.STAT\_CAUSE\_CODE ORDER BY Count(*) DESC LIMIT 1} \\
        \midrule
        \textbf{Question}  & What is the total area that has been burned until now?\\
        \textbf{no desc.} & \sql{SELECT Sum(*) FROM Fires} \\
        \textbf{desc.} & \sql{SELECT Sum(Fires.FIRE\_SIZE) FROM Fires} \\
        \bottomrule
        \toprule
        \textbf{Database \sql{Pesticide}} & \textbf{Column Descriptions}\\
        \sql{origin}  & Code indicating sample origin (1=U.S. 2=imported 3=unknown)\\
        \sql{country}  & Country of origin if the sample was imported\\
        \midrule
        \textbf{Question} & How many samples come from other countries? \\
        \textbf{no desc.} & \sql{SELECT sampledata15.country FROM sampledata15} \\
        \textbf{desc.}  & \sql{SELECT Count(*) FROM sampledata15 WHERE sampledata15.origin =} '$\dottedsquare$' \\
        %\hline
        % \textbf{NL}  & What are the salaries of players who have ever entered hall of fame? \\
        % no desc & SELECT salary.salary FROM hall\_of\_fame JOIN salary ON salary.player\_id = hall\_of\_fame.player\_id \\
        % & JOIN player ON hall\_of\_fame.player\_id = player.player\_id WHERE player.name\_first = 'terminal' \\
        % & AND salary.year = 'terminal' AND salary.year = 'terminal' \\
        %desc & SELECT salary.salary FROM hall\_of\_fame JOIN salary ON hall\_of\_fame.player\_id = salary.player\_id \\
        %& WHERE hall\_of\_fame.inducted = 'terminal' \\
        %\hline
        % \textbf{NL}  & Find me top 10 albums ranked by their popularity. \\
        %  no desc & SELECT torrents.totalSnatched FROM torrents WHERE torrents.totalSnatched = 'terminal' \\
        % & ORDER BY torrents.totalSnatched Desc LIMIT 1 \\
        % desc & SELECT torrents.groupName FROM torrents WHERE torrents.releaseType = 'terminal' \\
        % & ORDER BY torrents.totalSnatched Desc LIMIT 1 \\
        % \textbf{NL} & Which reactor type has the largest average capacity?  \\
        %\textbf{no desc} & SELECT \underline{nuclear\_power\_plants.ReactorModel} FROM nuclear\_power\_plants \\
        %& GROUP BY \underline{nuclear\_power\_plants.ReactorModel} \\
        %& ORDER BY Avg(nuclear\_power\_plants.Capacity) Desc LIMIT 1 \\
        %\textbf{desc} & SELECT \underline{nuclear\_power\_plants.ReactorType} FROM nuclear\_power\_plants \\
        %& GROUP BY \underline{nuclear\_power\_plants.ReactorType} \\
        %& ORDER BY Avg(nuclear\_power\_plants.Capacity) Desc LIMIT 1 \\
        \bottomrule
    \end{tabular}}
\end{table*}

\begin{table*}[t]
    \small
    \centering
    \caption{Distribution of error types in each domain over 10 randomly-selected erroneous examples.}
    \label{tab:error}
    % \vspace{-0.5\baselineskip}
    \begin{tabular}{lccccccccr}
        \toprule
        \textbf{Error Types}  & Nuclear & Crime & Pesticide & MathScore & Baseball & Fires & WhatCD & Soccer  & \% \\
        \midrule
        Entity-column matching & 0  & 2 & 2 & 3 & 0 & 2 & 0  & 0  &  15.00\% \\
        Incorrect Final Column & 3 & 2 & 5 & 3 & 4 & 4 & 4 & 2 & 33.75\%\\
        Missing Constraint  & 5 & 3 & 3 & 1 & 5 & 5 & 2 & 2 & 32.50\%\\
        Incorrect Constraint  & 4 & 2 & 2 & 2 & 6 & 0 & 7 & 2 & 31.25\%\\
        Understanding Error  &  0 & 1 & 0 & 4 & 0 & 1 & 2 & 3 & 13.75\% \\
        Ambiguous Columns & 0  & 2 & 2 & 0 & 1 & 1 & 0 & 0 & 7.50\%\\
        Equivalent  & 1 & 0 & 2 & 0 & 0 & 0 & 0 & 0 & 3.75\%\\
        \bottomrule
    \end{tabular}
\end{table*}

\subsubsection{Leveraging Database Documentation}
The database schemas in \systemname are obscure, making the task difficult without leveraging the database
documentation. We consider only the column descriptions, but other portions of the documentation may prove useful in
future work. The best approach for incorporating column descriptions into a text-to-SQL model is model-specific. RAT-SQL
makes use of relations between question tokens and schema terms to assist with \textit{schema-linking}. We extend the
same functionality to column descriptions by appending the column descriptions to the column names (separated by a
period) and recomputing matching relations. The concatenated column name is also presented to the transformer encoder
for schema encoding.
% \footnote{Not every column in \systemname has a description in the documentation. For others, we construct artificial descriptions as described in the next paragraph.}

Simply adding these descriptions results in mismatch between the training set (Spider) which does not have
descriptions, and the evaluation set (\systemname) which does.
% \matt{with column descriptions, there are significantly more partial-match relation edges between the question and schema in RAT-SQL, so...}
To alleviate it, we first augment the schemas in Spider with artificial descriptions.
For column $c$ of table $t$, the description for $c$ is ``\textit{the $c$ of the $t$}''.
We then retrain RAT-SQL on Spider with these artificial descriptions.

% Note that the model still has seen only artificial descriptions at train time.
Since the artificial descriptions simply restate information from the schema, the model may not learn to
leverage them for any further information about schema linking and simply treat them as noise.
Therefore, we also evaluate RAT-SQL adapted to the general domain of \systemname so that it (a) experiences useful
descriptions and (b) adapts to the language distribution of \systemname.
We evaluate the benefits of this \textit{adaptation} using leave-one-out: for each domain in \systemname, we fine-tune the model
on all other domains except for the target (with the same fine-tuning parameters as for few-shot learning).
Adapting in this way is predictive of the performance of a novel domain with similar characteristics.

As with the other few-shot results, the model is then fine-tuned on the few examples of target domain data.
\textit{Adaptation} and \textit{fine-tuning} are two separate training processes. \textit{Adaptation} is meant to adapt to the real-world distribution. 
\textit{Fine-tuning} is meant to adjust for in-domain knowledge. The most effective setting for a target database in our experiments is to conduct \textit{adaptation} first, followed by \textit{fine-tuning}.

\Cref{tab:indomain} (row (d)) shows the results.
Using column descriptions in the context of adaptation increases model accuracy from 17.96\% to 26.77\%.
Ablations show that adaptation and descriptions each contribute
approximately half of this gain (row (c)).
Descriptions provide no benefit without adaptation (row (b)), likely due to the train-test
mismatch between artificial descriptions and real ones.
Without any artificial descriptions, accuracy drops even further so they are critical to leveraging in-domain knowledge.
Overall, incorporating in-domain data (\ie a few-shot setting and database documentation) nearly doubles model accuracy from 13.56\% to 26.77\% on  \systemname.

\begin{table*}[t]
    \small
    \centering
    \caption{Statistics of each database in \systemname.}
        % All numbers are the exact match accuracy as defined by Spider evaluation metric.
    \label{tab:DBanalysis}
    \vspace{-0.5\baselineskip}
    \begin{tabular}{lcccccccc}
        \toprule
          & Nuclear & Crime & Pesticide & MathScore & Baseball & Fires & WhatCD & Soccer  \\
        \midrule
        %Hardness & \textbf{2.59} & \textbf{2.94} & 2.08 & \textbf{2.68} & \textbf{2.59} & 2.2 & 2.57 & 1.75 \\
        %Avg. Q len & 9.31 & 9.33 & 9.46 & \textbf{12.6} & \textbf{10.28} & \textbf{10.64} & 7.95 &  9.88 \\
        \#Tables & 1 & 1 & 2 & 3 & 5 & 1 & 2 & 2 \\
        \#Columns & 15 & 6 & 34 & 15 & 44 & 19 & 10 & 37 \\ 
        \#Fine-tuning Examples & 10 & 9 & 16 & 9 & 12 & 12 & 13 & 6 \\
        \#Test Examples & 22 & 18 & 34 & 19 & 27 & 25 & 28 & 12 \\
        \bottomrule
    \end{tabular}
\end{table*}

\begin{table*}[t]
    \small
    \centering
    \caption{The most common error types of our best model and their representative examples.}
    \label{tab:error_example}
    % \vspace{-0.5\baselineskip}
    \begin{tabular}{lp{14.2cm}}
        \toprule
        \multicolumn{2}{c}{\bfseries 33.75\%:  Incorrect Final Column} \\
        Question  &  What is the latitudinal band that is most likely to experience wildfires in the USA?\\
        Predicted &  \sql{SELECT STAT\_CAUSE\_DESCR FROM Fires GROUP BY STAT\_CAUSE\_DESCR ORDER BY Count(*) Desc LIMIT 1}\\
        Gold & \sql{SELECT LATITUDE FROM Fires GROUP BY LATITUDE ORDER BY count(*) DESC LIMIT 1} \\
        \midrule
        \multicolumn{2}{c}{\bfseries 32.5\%:  Missing Constraint} \\
        Question  &  How many nuclear power plants are in preparation to be used in Japan? \\
        Predicted & \sql{SELECT Count(*) FROM nuclear\_power\_plants WHERE Country =} '$\dottedsquare$' \\
        Gold & \sql{SELECT count(*) FROM nuclear\_power\_plants WHERE Country = "Japan" AND Status = "Under Construction"} \\
        \midrule
        \multicolumn{2}{c}{\bfseries 31.25\%:  Incorrect Constraint} \\
        Question  &  Which state gets the highest revenue?\\
        Predicted &  \sql{SELECT NDECoreExcel\_Math\_Grade8.state FROM FINREV\_FED\_17 JOIN NDECoreExcel\_Math\_Grade8 GROUP BY NDECoreExcel\_Math\_Grade8.state ORDER BY Sum(FINREV\_FED\_17.t\_fed\_rev) Asc}\\
        Gold &   \sql{SELECT T2.state FROM FINREV\_FED\_KEY\_17 as T2 JOIN FINREV\_FED\_17 as T1 ON} \\ & \sql{T1.state\_code = T2.state\_code GROUP BY T2.state ORDER BY sum(t\_fed\_rev) DESC LIMIT 1}\\
        \midrule
        \multicolumn{2}{c}{\bfseries 15\%:  Entity-column matching} \\
        Question  & Which type of crime happens the most in Salford? \\
        Predicted &  \sql{SELECT Type FROM GreaterManchesterCrime WHERE Location LIKE} '$\dottedsquare$' \sql{GROUP BY Type ORDER BY Count(*) Desc LIMIT 1}\\
        Gold &  \sql{SELECT Type FROM GreaterManchesterCrime WHERE LSOA LIKE "\%Salford\%" GROUP BY Type ORDER BY count(*) DESC LIMIT 1} \\
        \midrule
        \multicolumn{2}{c}{\bfseries 13.75\%:  Understanding Error} \\
        Question  & How many downloads of ep and album respectively? \\
        Predicted &  \sql{SELECT Sum(totalSnatched), Sum(totalSnatched) FROM torrents WHERE releaseType =} '$\dottedsquare$' \\
        Gold &   \sql{SELECT sum(totalSnatched) FROM torrents WHERE releaseType = "ep"   UNION} \\
        & \sql{SELECT sum(totalSnatched) FROM torrents WHERE releaseType = "album"}\\
        \bottomrule
\end{tabular}
\end{table*}

\subsection{Column Normalization}
One of the major challenges in \systemname is that column names are often obscure or abbreviated. A natural question is whether this creates difficulty because the model struggles to understand the meaning of a column or because it leads to a low overlap between question and column terms. In an attempt to tease these factors apart, we created a \textit{normalized} version of \systemname by replacing the obscure column names with normalized column names such as one might find in the Spider dataset. This was done manually using column descriptions to help clarify each column and without introducing any extra knowledge into the column names except for the expansion of abbreviations (\eg \texttt{t\_fed\_rev}~$\to$ \texttt{total federal revenue}).

In Table~\ref{tab:normalized} we give the results of evaluation on the normalized \systemname, following the same setup as Table~\ref{tab:indomain}. Normalization provides a significant boost in performance (row~(c) vs. row~(a)).
The trend is similar to Table~\ref{tab:indomain}. Without adaptation, models with descriptions are not better than those without (row~(b) vs. row~(a), row~(d) vs. row~(c)). After adaptation, the train-test mismatch is partly mitigated and the performance improves (row~(f) vs. row~(e), row~(h) vs. row~(g)). Normalization and descriptions provide complementary knowledge augmentation, jointly improving accuracy by 5\% (row~(h) vs. row~(e)), more than either alone.

Normalization helps clarify the obscure column names of \systemname. However, the other challenges such as low column mention percentage and in-domain schema conventions still leave significant room for improvement. We provide the full experimental results on normalized tables in the Appendix.

\subsection{Error Analysis}
% In this section, we qualitatively analyze improvements made by augmenting descriptions, and analyze outstanding issues.

\Cref{tab:correct} shows examples of improvements due to descriptions.
First, column descriptions help the parser correctly identify columns to select.
For instance, it chooses \sql{STAT_CAUSE_CODE} over \sql{STAT_CAUSE_DESCR} when asked for ``the most common cause of the
fire (code)''.
Second, they clarify necessary constraints.
For instance, when asked ``how many samples come from other countries?'', the parser chooses the correct \sql{origin}
column rather than superficially-matching \sql{country} in the clause
\sql{WHERE sampledata15.origin = "2"}.

\Cref{tab:error} shows a distribution of error types in \systemname using 10 randomly-selected erroneous
predictions for each domain.
The error categories mostly follow \citet{suhr2020exploring}, modulo (a) removing unobserved categories, (b) separating
semantically equivalent predictions into their own ``Equivalent'' category, and (c) categorizing significant structural
errors as ``Understanding Errors''. We also provide more characteristics of each database in Table~\ref{tab:DBanalysis} in an attempt to understand the difference in performance across databases. Our model performs worst on the databases with the most columns (\textit{Pesticide}, \textit{Baseball} and \textit{Soccer}). The only database with lower accuracy is \textit{MathScore} which has multiple tables and a relatively small fine-tuning set.

The most common error types and their examples are summarized in~\Cref{tab:error_example}.
\textbf{(i)}~The most common type is ``Incorrect Final Column'' (33.75\%),
illustrating the difficulty of schema linking in \systemname even with documentation and fine-tuning.
\textbf{(ii)}~32.5\% of the errors are in ``Missing Constraints''.
In \systemname questions, users sometimes use implications instead of directly mentioning the desired constraint, \eg
``in preparation'' for \sql{Status = "Under Construction"}.
\textbf{(iii)}~31.25\% of the errors are in ``Incorrect Constraint'', \eg failing to parse ``highest'' into the
top-1 result in descending order.
\textbf{(iv)}~15\% of the errors are in ``Entity-column matching'', \eg aligning ``Salford'' to \sql{Location}
rather than \sql{LSOA}.
This illustrates the difficulty of \textit{value linking}, partly mitigated by value descriptions for categorical
columns in the database documentation.

\section{Conclusion \& Future Work}
\systemname provides two resources to facilitate real-world
applications of text-to-SQL parsing.
First, it encourages an evaluation regime that bridges the gap between academic and industrial settings, leveraging
in-domain knowledge and more realistic database distribution.
We encourage adopting this regime for established text-to-SQL benchmarks.
Second, it is a new dataset of more realistic databases and questions, presenting a challenge to
state-of-the-art parsers.
Despite the addition of domain knowledge in the form of database documentation,
our baselines reach only 26.77\% accuracy, struggling to generalize to harder questions.
We hope that better use of documentation and new modeling and domain adaptation techniques will help further advance
state of the art.
The \systemname dataset is available at \mbox{\url{https://aka.ms/KaggleDBQA}}.

\section*{Ethical Considerations}

%Ensure that your dataset has been collected in a manner which is consistent with the terms of use of any sources and the intellectual property and privacy rights of the original authors of the texts. The fact that a dataset has already been used previously does not necessarily make it acceptable. If appropriate, check that informed consent was signed. Your institution may have a data officer or a review board charged with helping researchers navigate these issues and we encourage you to reach out to them for assistance.

%Detail the dataset collection process and conditions. If manual work was involved, describe measures taken to ensure that crowd workers or other annotators were fairly compensated and how fair compensation was determined.

\paragraph{Dataset Collection}
The data collection process was pre-approved by IRB.
Each annotator agreed to a consent form before having access to the labeling task. 
Each annotator was rewarded with a \$20 e-gift card for the approximately one hour of their time.
The authors of this paper acted as the SQL annotators and incurred no additional compensation. The databases collected for \systemname were individually reviewed to ensure they were properly licensed for re-distribution.
For other details of dataset construction, please refer to~\Cref{sec:dataset}.

% Does the paper describe how participants’ privacy rights were respected in the data collection process?
Aside from email addresses, no personal information of annotators was collected during our study. Email addresses were not shared and were promptly deleted after compensation had been provided. The association between annotator and annotation was deleted before any analysis or distribution was conducted.

%Please indicate if your project was approved by an IRB (institutional review board). (For a discussion of IRBs in the US context, see https://www.apa.org/advocacy/research/defending-research/review-boards)

%Describe the characteristics of the dataset in enough detail for a reader to understand which speaker populations the technology could be expected to work for. (For suggestions of what kind of information to include, see Bender and Friedman 2018, Mitchell et al 2019 and Gebru et al 2018.)

%Finally, describe the steps taken to ensure that potential problems with the quality of the dataset do not create additional risks.
%Also, check https://2021.naacl.org/ethics/review-questions/

\paragraph{Language Distribution}
\systemname only includes question annotations and databases in English, thus evaluating multi-lingual text-to-SQL
models on it will require translation.
The set of annotators included both native and second-language speakers of English, all fluent.

\paragraph{Usage of DBQA Technology}
Our goal with \systemname is to encourage the development of DBQA that will work in real-world settings. The actual deployment of a text-to-SQL parser must be conducted with appropriate safeguards in place to ensure users understand that the answers may be incorrect, especially if those answers are to be used in decision making.

%\section*{Acknowledgments}

%The acknowledgments should go immediately before the references. Do not number the acknowledgments section.
%\textbf{Do not include this section when submitting your paper for review.}

\bibliographystyle{acl_natbib}
\bibliography{acl2021}

\clearpage
\begin{table*}[t]
    \centering
    \small
    \caption{Evaluation results on \systemname using 100\% of the evaluation data. All numbers are the exact match accuracy evaluated by the Spider official scripts. Here we report the average score of three runs with different random seeds.}
    \label{tab:outofdomain}
     \adjustbox{max width=\textwidth}{\begin{tabular}{lccccccccc}
        \toprule
        Models  & Nuclear & Crime & Pesticide & MathScore & Baseball & Fires & WhatCD & Soccer & Avg  \\
        \midrule
        %EditSQL & & & & & & & &  &  11.03\\
        RATSQL &  22.91 &  23.45  &   8.00  &     0.00  &  11.11  & 25.22  & 4.76 & 11.11  & 13.32   \\
        \quad \textit{w.} desc  & 21.87 &   20.98    &    9.99  &   0.00  & 11.11 &  18.01 &  6.50 &  11.11 &  12.44 \\
        \quad \textit{w.} adaptation   & 20.83  & 33.33 &  12.66 & 3.57 & 11.11 &  24.32 & 8.93 &  12.96 & 15.96\\
        \quad \textit{w.} desc \textit{+} adaptation  & 29.16 &   25.88    &    18.00  &   3.57  & 16.23 &  30.62 & 10.53 &  12.96 &  18.37 \\
        \bottomrule
    \end{tabular}}
\end{table*}

\begin{table*}[t]
    \small
    \centering
    \caption{The original user question distribution. This reflects the natural information need from users.}
    \label{tab:question_distribution}
    \begin{tabular}{lll}
        \toprule
        \textbf{Question Types}  & \# &  Example \\
        \midrule
        Yes/No &   \multirow{2}{*}{51} & Has there been a recent surge in violent crime in Manchester? \\
        Percentage &   &  What percentage of August crime detections resulted in prosecution of a suspect? \\
        \midrule
        Time-related & 46 & Divide the day into 3 slots (6am to 4pm, 4pm to 11pm, 11pm to 6am), \\
                     & & which has the highest amount of crime conducted per hour? \\
        SQL-unexpressible &  31 & Which states had the highest percentage change in average scores \\
                          & & over the last few years? \\
        SQL-expressible & 272 &  Which LSOA has had the most instances of bicycle theft this month?  \\
        \bottomrule
    \end{tabular}
\end{table*}

\appendix
%\appendixpage
\section{Appendix}
\subsection{Evaluation on Full Testing Data}
We show the zero shot testing and out-of-domain adaptation results in~\Cref{tab:outofdomain}.
In contrast to~\Cref{tab:indomain}, they are evaluated using the full set of testing data.

\subsection{Details of Dataset Construction}
\subsubsection{Example Page of User Instructions}
For each user, we show two different HTML files that contain different instructions of the task, database overview, table name(s), column descriptions, ten sampled rows of the database content.

\subsubsection{Question Types}
\label{sec:question_types}
Question annotators were allowed to write any type of question without restriction. While this represents a natural distribution of questions one might expect to encounter in a realistic setting, some types do not appear in the Spider training set and thus pose particular difficulty with current text-to-SQL systems. We remove these from the official evaluation but still include them in the dataset for future work on these types of questions. Table~\ref{tab:question_distribution} summarizes the distribution over these types of questions.
\subsubsection{SQL annotation Guidelines}
\label{sec:annotation}
%We paraphrase the non-deterministic questions (e.g., "Has there been a recent surge in violent crime in Manchester?") and ambiguous questions (e.g., During what time of day are there most crimes?).
%However, different from the previous datasets, we keep the questions that require certain extent of common sense knowledge(e.g., Which country is the safest in terms of the pesticide concentration found in imported foods?).
We also establish few guidelines and follow them throughout the annotation process:
\begin{enumerate}[left=0pt]
    \item If the referred column is
        categorical, use "=" operator with the value from the database (e.g., \textit{Where is the area with the largest
            number of sexual offenses crime events?} \textrightarrow\ \sql{SELECT Location FROM GreaterManchesterCrime WHERE Type = "Violence and sexual offences" GROUP BY Location ORDER BY count(*) DESC LIMIT 1}).
            If it is free-form text use
            "LIKE" operator with a term from the question (e.g., \textit{What were the closing odds for a draw in matches with VfB Stuttgart?} \textrightarrow\ \sql{SELECT DRAW_CLOSING FROM betfront WHERE MATCH LIKE "\%VfB Stuttgart\%"}).
              %"LIKE" operator with a term from the question (e.g., \textit{Which date is the article containing "captain
              %  america operation zero point" being published?} \textrightarrow\ \sql{SELECT T2.date FROM quotes as T1 JOIN articles as T2 ON T1.article\_id = T2.article\_id WHERE T1.phrase LIKE "\%captain america operation zero point\%"}).    
            \item Sometimes ID columns are paired with their name realizations (e.g., \sql{state_code} and \sql{state})
        We choose to return ID whenever users do not explicitly ask for the name realizations.
    \item Duplicate rows can sometimes yield an incorrect result.
        However, it is not possible for models to know in advance unless they encode database content.
        So we use the \sql{DISTINCT} operator when necessary to return the correct answer or it is explicitly asked for
        by the user (e.g., \textit{What are titles for each unique entry?}).
\end{enumerate}

\begin{table*}[t]
    \small
    \centering
    \caption{Exact match accuracy and standard error on schema-normalized \systemname, average of three runs with different random seeds.}
        % All numbers are the exact match accuracy as defined by Spider evaluation metric.
    \label{tab:indomain_normalized}
    \vspace{-0.5\baselineskip}
     \adjustbox{max width=\textwidth}{\begin{tabular}{lccccccccc}
        \toprule
        \multicolumn{10}{c}{With \textit{fine-tuning}} \\
        \midrule
        Models  & Nuclear & Crime & Pesticide & MathScore & Baseball & Fires & WhatCD & Soccer & Avg  \\
        \midrule
        (a) RAT-SQL  & 25.75 & 44.44 & 23.52 & 7.01  & 19.74  & 33.33  & 22.61 & 8.33  &  $23.09  \pm 0.9\% $\\
        (b) \quad \textit{w.} desc & 25.75  &  40.73 &  19.60  &  3.50 & 20.98 & 28.00 & 25.00 & 8.33 & $21.48 \pm 1.0\%$\\
        (c) \quad \textit{w.} \textit{adaptation}  & 30.30  & 46.29 & 19.60 & 12.27 & 19.74 & 41.33 & 21.42 & 13.88 & $25.60 \pm 0.9\%$ \\
        (d) \quad \textit{w.} desc \textit{+} \textit{adaptation}  & 33.33 & 49.99 & 28.43 & 8.76 & 22.21 & 37.33 & 26.18 & 16.44 & \textbf{$27.86  \pm 0.7\%$} \\
        \midrule
        \multicolumn{10}{c}{Without \textit{fine-tuning}} \\
        \midrule
        Models  & Nuclear & Crime & Pesticide & MathScore & Baseball & Fires & WhatCD & Soccer & Avg  \\
        \midrule
        (e) RAT-SQL   & 30.29 & 35.18 & 15.68 & 0.05  & 12.34  & 22.66  & 5.95 & 25.00  &  $19.04  \pm 0.6\% $\\
        (f) \quad \textit{w.} desc   & 24.23 & 25.92 & 13.72 & 0.00  & 0.08  & 13.33  & 0.07 & 13.87  &  $13.35  \pm 0.9\% $\\
        (g) \quad \textit{w.} \textit{adaptation}   & 25.75  & 40.73 &  21.56 & 14.02 & 14.81 &  25.33 & 10.69 &  25.00 & $22.23 \pm 0.7\%$\\
        (h) \quad \textit{w.} desc \textit{+} \textit{adaptation}  & 34.84 &   37.03   &   23.52  &   8.76  & 18.51 &  24.00 &  16.66 &  21.96 &  $23.16  \pm 0.5\%$\\
        \bottomrule
    \end{tabular}}
\end{table*}

\subsection{Implementation Details}
\label{sec:implementation_details}
For all our experiments we use the RAT-SQL official implementation and the pre-trained BERT-Large from Google.~\footnote{We use the BERT-Large, Uncased (Whole Word Masking) model from \url{https://storage.googleapis.com/bert_models/2019_05_30/wwm_uncased_L-24_H-1024_A-16.zip}}
We follow the original settings to get the pre-fine-tuned/pre-adapted models.
For adaptation and fine-tuning, we decrease the learning rate of BERT parameters by 50 times to 6e-8
to avoid overfitting.
We keep the learning rate of non-BERT parameters the same at 7.44e-4.
We also increase the dropout rate of the transformers from 0.1 to 0.3 to provide further regularization.

%\begin{figure*}[p]
%    \centering
%    \includegraphics[width=\linewidth, trim={40 0 40 0}, clip]{fig/html.pdf}
%    \caption{An example instruction page for the annotation task.}
%    \label{fig:html}
%\end{figure*}

\end{document}